\documentclass[letterpaper, 10 pt, conference]{ieeeconf}
\IEEEoverridecommandlockouts
\usepackage{epsfig} 
\usepackage{amsmath}
\usepackage{multirow}
\usepackage{tabularx}
\usepackage{color}
\usepackage{algorithm}
\usepackage{algorithmicx}
\usepackage{algpseudocode}
\usepackage{bm}
\usepackage{amsfonts}
\usepackage{graphics}
\usepackage{url}
\usepackage{siunitx}
\usepackage{mathrsfs}
\usepackage{soul}
\usepackage{flushend}
\usepackage{caption}
\usepackage{subcaption}
\usepackage[dvipsnames]{xcolor}
\usepackage{hyperref}
\hypersetup{backref=true,
pagebackref=true,
hyperindex=true,
colorlinks=true,
breaklinks=true,
urlcolor= black,
linkcolor= blue,
bookmarks=true,
                    bookmarksopen=false,
                    citecolor=black,
                    linkcolor=black,
                    filecolor=black,
                    citecolor=blue,
                    linkbordercolor=blue
}

\algtext*{EndWhile}
\algtext*{EndIf}
\algtext*{EndFor}

\soulregister{\cite}7
\soulregister{\citep}7
\soulregister{\citet}7
\soulregister{\ref}7
\soulregister{\pageref}7
\soulregister{\uppercase\expandafter}7
\sethlcolor{yellow}

\newcommand{\deletefig}[1]{{\bgroup\markoverwith{\textcolor{red}{\rule[2.5ex]{2pt}{2.0pt}}}\ULon{#1}}}

\newcommand{\delete}[1]{}

\title{ \LARGE \bf
Decentralized Multi-Agent Trajectory Planning in Dynamic Environments with Spatiotemporal Occupancy Grid Maps
}

\author{Siyuan Wu, Gang Chen, Moji Shi, and Javier Alonso-Mora
\thanks{All authors are with the Department of Cognitive Robotics (CoR), Delft University of Technology, 2628CD Delft, The Netherlands.
	{\tt\small \{s.wu-14; m.shi-5\}@student.tudelft.nl; \{g.chen-5; j.alonsomora\}@tudelft.nl}
This work is funded in part by the European Union (ERC, INTERACT, 101041863). Views and opinions expressed are however those of the author(s) only and do not necessarily reflect those of the European Union or the European Research Council Executive Agency. Neither the European Union nor the granting authority can be held responsible for them.
}
}

\begin{document}
\maketitle
\thispagestyle{empty}
\pagestyle{empty}

\begin{abstract}
This paper proposes a decentralized trajectory planning framework for the collision avoidance problem of multiple micro aerial vehicles (MAVs) in environments with static and dynamic obstacles. The framework utilizes spatiotemporal occupancy grid maps (SOGM), which forecast the occupancy status of neighboring space in the near future, as the environment representation. Based on this representation, we extend the kinodynamic A* and the corridor-constrained trajectory optimization algorithms to efficiently tackle static and dynamic obstacles with arbitrary shapes. Collision avoidance between communicating robots is integrated by sharing planned trajectories and projecting them onto the SOGM. The simulation results show that our method achieves competitive performance against state-of-the-art methods in dynamic environments with different numbers and shapes of obstacles. Finally, the proposed method is validated in real experiments.
\end{abstract}


\section{Introduction} \label{Section : Introduction}

Decentralized multi-agent trajectory planning (MATP) problem in dynamic environments remains a challenging problem for years.
To solve this problem, the robot must avoid collisions with static and dynamic obstacles during planning, as well as with other MAVs.
A common approach to planning in dynamic environments inherits the pipeline used in static environments with modifications to account for dynamic obstacles \cite{wang2021Autonomous,9359513}.
These methods rely on an Occupancy Grid Map (OGM) to represent static obstacles and generate a collision-free trajectory for each MAV \cite{zhou2022Swarm}.
The trajectory is then further optimized to avoid dynamic obstacles and other robots, which are tracked and represented separately with fixed-shape models, e.g., ellipsoids \cite{zhou2022Swarm, zhu2019ChanceConstrained}, columns \cite{xu2022DPMPCPlanner} and axis-aligned bounding boxes \cite{tordesillas2022MADER}.

However, those methods require distinct pipelines for static and dynamic obstacles and other robots, often leading to trajectories that avoid dynamic obstacles but conflict with static ones or other robots.
To address this issue, we use Spatiotemporal Occupancy Grid Maps (SOGM) \cite{Learning2022Tomas,OursMap} to capture both the current and future occupancy status of arbitrary-shaped static and dynamic obstacles in the environment.
Moreover, we incorporate the occupied space of other robots within the same time window to achieve a unified representation of both obstacles and other robots.
Based on this representation, we employ spatio-temporal safety corridors to identify areas free from obstacles and other robots, effectively solving the MATP in dynamic environments.

\begin{figure}[t]
	\centering
	\includegraphics[width=1.0\linewidth]{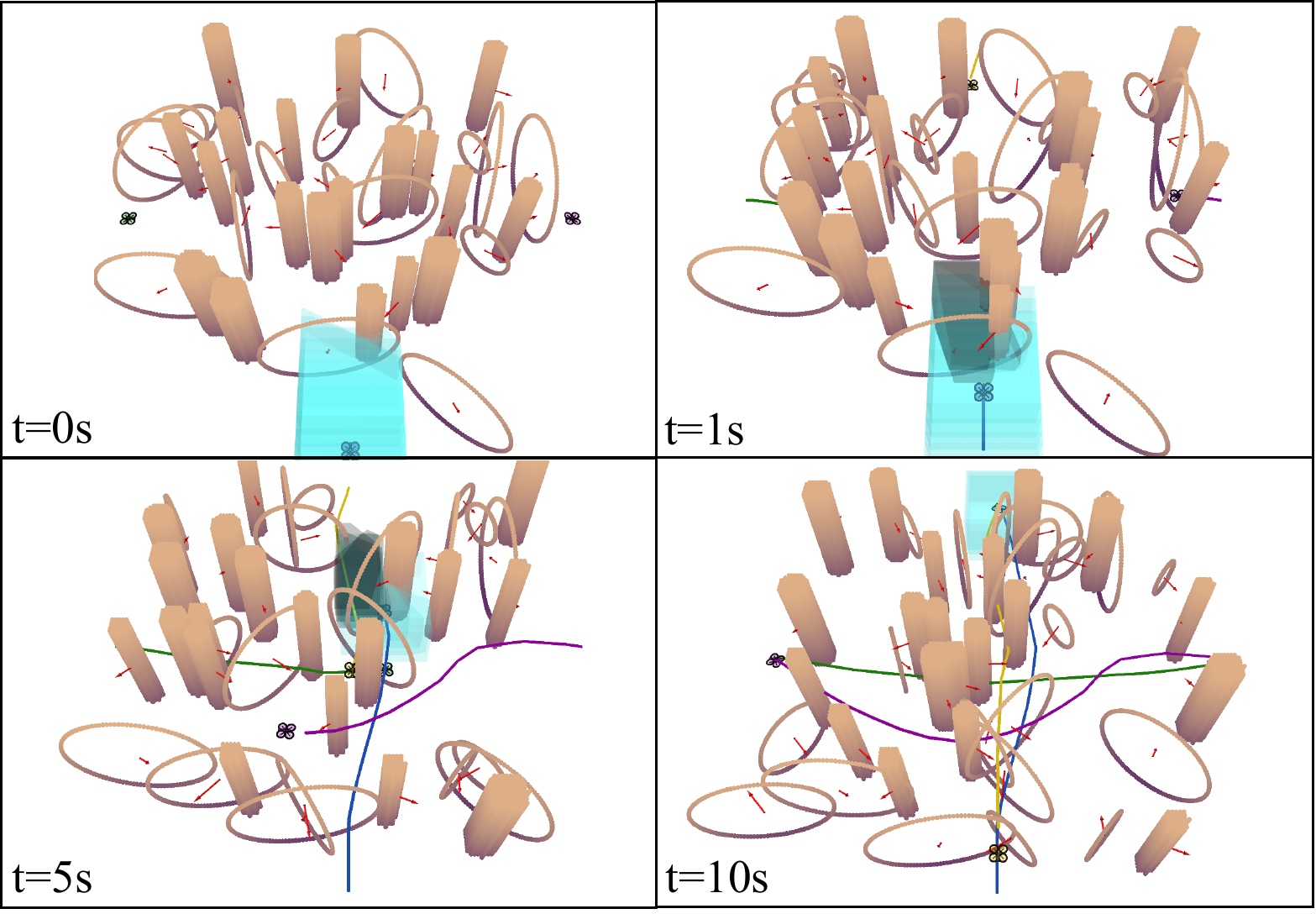}
\caption{4 robots navigate in a dynamic environment with 20 columns and 20 circles.
All robots perform independent planning and coordinate with trajectory sharing.
	The spatio-temporal corridors are indicated in blue, and obstacle velocities are marked by red arrows.
	}
	\label{fig:header}
\end{figure}

\begin{figure*}[ht]
	\centering
	\includegraphics[width=1.0\linewidth]{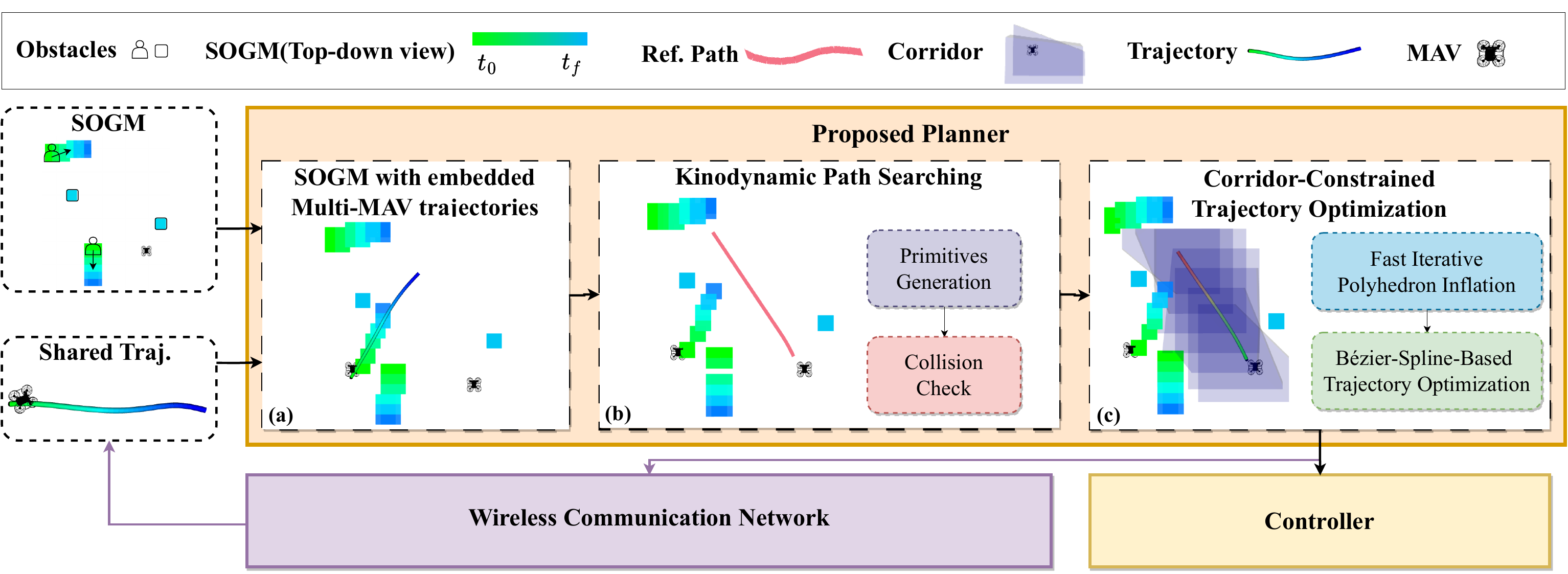}
	\caption{Overview of the proposed framework for decentralized multi-robot trajectory planning.}
	\label{fig:overview}
\end{figure*}

In this work, we assume that the SOGM of each robot is given, and extend a corridor-based trajectory optimization method to MATP in dynamic environments.
First, a kinodynamically feasible path is searched within the map, and spatio-temporal corridors are generated to maximize the free space.
Next, trajectories are generated by solving a minimum jerk trajectory optimization problem with the spatio-temporal corridors as constraints.
To prevent collisions between robots, other robots are treated as moving obstacles with known shapes, and their trajectories are shared via a communication network and represented simultaneously in the SOGM.
For dynamic obstacles, such as robots that do not communicate, we employ a constant velocity prediction.
Compared to existing MATP methods, our planner is capable of safely navigating multiple MAVs in complex environments with arbitrarily shaped static and dynamic obstacles.
Our contributions are summarized as:
\begin{enumerate}
	\item A decentralized multi-robot collision avoidance framework that extends the corridor-based methods used in static environments to dynamic environments.
	\item An efficient planning approach that generates safe trajectories in the SOGM and is capable of avoiding static and dynamic obstacles with arbitrary shapes.
\item A paradigm that integrates shared trajectories into occupancy grids to build corridors free from obstacles and other robots.
\end{enumerate}

We will release our code for the community’s reference.

\section{Related Work}  \label{sec:related-works}

\subsection{MAV Planning in Dynamic Environment}
MAV trajectory planning methods in dynamic environments can be divided into two categories, depending on whether they consider static and dynamic obstacles separately or jointly.
The first category \cite{zhou2022Swarm,xu2022DPMPCPlanner,hou2022Enhanced} usually employs an OGM \cite{hornung2013octomap, Esperance2014Safety} to model static obstacles, taking advantage of its ability to represent arbitrary 3D shapes \cite{lavalle2006Planning}.
However, dynamic obstacles are separately represented using a fixed-shape model with predicted trajectories, including ellipsoids \cite{lin2020Robust,zhu2019ChanceConstrained,xu2022DPMPCPlanner}, spheres \cite{Mina2017Robust}, or bounding boxes \cite{tordesillas2022MADER}.
Their future trajectories are normally predicted by polynomial fitting methods \cite{gao2017Quadrotor} or the constant velocity model \cite{chen2022Realtime}.
This representation of dynamic obstacles narrows the free space and increases the difficulty of trajectory generation by introducing non-convex collision avoidance constraints.
Static and dynamic obstacles can be represented jointly.
\cite{10034468, 9359513} represents static and dynamic obstacles with a particle-based occupancy map \cite{OursMap} and uses a sampling-based method to find a trajectory that encounters fewer particles.
Further in \cite{chen2023RAST}, risk-aware spatio-temporal corridors are generated on the map for trajectory optimization.
The trajectory generation method in \cite{chen2023RAST} is limited to cubic corridors, which can be overly conservative, especially in environments with arbitrarily shaped obstacles.
Inspired by \cite{chen2023RAST}, this work aims to extend these corridor-based methods to solve the MATP problem by addressing the aforementioned limitations.


\subsection{Decentralized Multi-Agent Trajectory Planning}
Multi-agent trajectory planning in complex environments is a challenging task in aerial robotics.  
A few methods have been proposed to solve this problem in a decentralized manner for scalability and robustness \cite{honig2018Trajectory,zhu2019BUAVC}.
In these methods, robots usually coordinate with each other by sharing their current states or planned trajectories.
With the shared states, collision-free trajectories can be generated by constructing Buffered Voronoi Cells \cite{zhu2019BUAVC} or distributed MPC \cite{luis2020Online}, which provide better safety guarantees but inhibit flight speed.
When using shared trajectories, it becomes possible to plan more agile routes. For example, MADER \cite{tordesillas2022MADER} and Robust-MADER \cite{kondo2023Robust} broadcast planned trajectories through a communication network, and propose a check-recheck paradigm to avoid collisions with other robots.
EGO-Swarm \cite{zhou2022Swarm} uses a reciprocal barrier cost to generate collision-free trajectories by solving unconstrained optimization problems.
In \cite{park2021Online}, shared trajectories are used to generate linear relative safety corridors as hard constraints.
In this work, we depict the shared trajectories in SOGM, facilitating simultaneous collision avoidance for robots and obstacles with the spatio-temporal safety corridors. 

\section{Methodology} \label{sec:methodology}

This section first briefly introduces how we use the SOGM in the MATP problem and describes the paradigm that integrates shared trajectories into SOGM (Sec. \ref{sec:meth:map}).
Then a kinodynamic path searching method designed for the SOGM is presented (Sec. \ref{sec:meth:path-finding}).
Finally, we describe the method that generates corridors from the reference path and introduce the trajectory optimization method to generate collision-free trajectories within these corridors (Sec. \ref{sec:meth:opt}).

\subsection{SOGM with Shared Trajectories Embedded} \label{sec:meth:map}

Various methods have been proposed to construct SOGMs from the point cloud, such as filtering-based methods \cite{OursMap} and learning-based methods \cite{Learning2022Tomas, mann2022Predicting}.
The resulting SOGM is a 4D OGM with spatial resolution $r_s$ and temporal resolution $r_\tau$.
In this work, we assume that the local SOGM denoted by $\mathbb{M}_k$ in the frame of robot $k \in \{ 1, 2, \dots, K \}$ with $T$-step predictions, which is generated by the perception algorithm in real-time (Fig.~\ref{fig:overview}a).
$\mathbb{M}_k(i)$ denotes the $i$-th frame of the SOGM $\mathbb{M}_k$, which is an OGM between $t_{i - 1}$ and $t_{i}$.

In the MATP problem, collision avoidance between non-cooperative obstacles and other cooperative robots is challenging.
The difficulty primarily arises from the distinct representations of the two types of obstacles: Non-cooperative obstacles are represented by the SOGM, while cooperative robots are characterized by their trajectories.
It requires two independent pipelines to achieve collision avoidance, which is inefficient and sometimes intractable because a collision-free trajectory optimized in one pipeline might be infeasible in another pipeline.
Therefore, the published trajectories of cooperative robots are projected in the SOGM by accounting for the occupied grids of the robots from one time step to the next.
Assume that robot $k$ receives shared trajectories $\boldsymbol{p}(t)_{k^\prime}$ and occupancy $\mathcal{O}_{k^\prime}$ from other robots $k^\prime \in \{ 1, 2, \dots, K \} \setminus k$.
Then the occupancy of robot $k^\prime$ traveled from $t_{i - 1}$ to $t_{i}$ is represented as
\begin{equation}
	\mathcal{O}_{k, k^\prime}[t_{i-1}, t_{i}] = \int_{t_{i-1}}^{t_i} \mathcal{O}_{k^\prime} \oplus \boldsymbol{p}(t)_{k^\prime} dt
\end{equation}
which the corresponding 3D spaces occupied in the SOGM $\mathbb{M}_{k}$ as $\mathcal{O}_{k^\prime}$ moves along trajectory $\boldsymbol{p}(t)_{k^\prime}$ in the local frame of robot $k$, as shown in Fig. \ref{fig:overview}a.
Therefore, the SOGM considering cooperative robots can be represented as
\begin{equation}
\mathcal{M}_{k}(i) = \left( \bigcup \limits_{k^\prime \in \{ 1, 2, \dots, K \} \setminus k} \mathcal{O}_{k, k^\prime}[t_{i-1}, t_{i}] \right) \cup \mathbb{M}_{k}(i).
\end{equation}
$\mathcal{M}_{k}(i)$ will be updated applying the recent $\boldsymbol{p}(t)_{k^\prime}$ once $\mathbb{M}_{k}(i)$ is updated.
	For the initial robot, $\mathcal{M}_{1}(i) = \mathbb{M}_{1}(i)$ for all $i \in \{ 1, 2, \dots, T \}$ since no other trajectories are published.



\subsection{Kinodynamic Path Searching in SOGMs} \label{sec:meth:path-finding}


We develop a path-searching method based on the kinodynamic algorithm A* \cite{zhou2019Robust} to find a kinodynamically feasible reference path (as shown in Fig.~\ref{fig:overview}b).
Similar to \cite{zhou2019Robust}, we select the double integrator as the approximate model and generate a set of primitives by discretizing the control inputs of the robot.
The primitive duration is set to the temporal resolution $r_\tau$ of the SOGM.
The cost function is defined as the cumulative square of control inputs over the time horizon.
The heuristic function is computed by applying Pontryagin's minimum principle \cite{zhou2019Robust} which yields the optimal control input to reach the goal state.
Since this function is both admissible and consistent, optimality is guaranteed.
In many path-searching methods, obstacles are pre-inflated by the robot radius to facilitate feasibility checks.
Due to the high dimension of the SOGM, inflating the obstacles with the robot radius directly on the map is not efficient.
Instead, we first inflate the primitive with the shape of the robot and then check if the inflated primitive collides with obstacles in the corresponding temporal frame $\mathcal{M}_{k}(i)$.

\subsection{Corridor Constrained Trajectory Generation} \label{sec:meth:opt}
We develop a trajectory optimization method for the MATP problem to generate collision-free trajectories constrained by spatio-temporal corridors inspired by \cite{chen2023RAST}.

\subsubsection{Spatiotemporal Corridor Generation}

Upon determining a collision-free reference path in the SOGM, our objective is to delineate the obstacle-free region along this path before proceeding with trajectory generation.
Due to the dynamic nature of the environment, the safety region of the path may change with time, so the static corridor introduced in \cite{liu2017Planning} is no longer applicable.
A natural approach is to locate the safety region at each temporal frame $\mathbb{M}_{k}(i)$ of the SOGM.
In this work, the spatio-temporal corridor is defined as a convex polyhedron $\mathcal{P} = \left\{ x \in \mathbb{R}^n | \boldsymbol{A}_{\mathcal{P}} x \leq \boldsymbol{b}_{\mathcal{P}} \right\}$ that encloses the safety region with the time window $[t_{i - 1}, t_{i}]$.
The time duration of each corridor is determined by the time resolution $r_\tau$ of the SOGM.

The problem of finding the obstacle-free convex corridor constraints for robot $R_{k}$ at temporal frame $\mathbb{M}_{k}(i)$ is formulated as the following optimization problem:
\begin{equation}
	\max_{ \boldsymbol{A}_{\mathcal{P}}, \boldsymbol{b}_{\mathcal{P}}} \text{vol}(\mathcal{P}), \text{ s.t. } R_{k} \subset \mathcal{P}, \mathcal{O}_{k, i} \subset \mathbb{M}_{k}(i) \setminus \mathcal{P},
\end{equation}
To efficiently solve the optimization problem, we apply Fast Iterative Region Inflation (FIRI) \cite{wang2022Geometrically}.
FIRI approximates the maximum volume polyhedron $\mathcal{P}$ by iteratively and monotonically increasing the volume of its inscribed ellipsoid.
It formulates the problem as an equivalent second-order conic programming (SOCP) problem to decrease the dimension of the problem and reduce the computational cost.
Thanks to its efficiency, we can generate a sequence of feasible spatio-temporal corridors at $\mathbb{M}_{k}$ in less than 5ms.


Given that the corridor produced by FIRI solely encompasses the obstacle-free region without accounting for the robot's size, it is essential to shrink the corridor's size to guarantee that the robot can navigate through.
This reduction is achieved by inwardly adjusting all the corridor edges by the robot's radius, $d$.
Following this adjustment, a linear programming check is employed to confirm the corridor remains feasible and that consecutive corridors are interconnected.

\subsubsection{Minimum Jerk Trajectory Optimization}

We employ the Bézier spline to parametrize the trajectory.
A Bézier spline is a $n$-th order piecewise polynomial defined as:
\begin{equation}
	\boldsymbol{p}_{j}(t) = \sum_{i=0}^n \boldsymbol{c}^i_j b_n^i(t - t_{j-1}) , \quad t \in [t_{j-1}, t_j),
\end{equation}
where $\boldsymbol{c}^i_j$ denotes the $i$-th control point at the $j$-th piece of Bézier spline.
$b_n^i(t) = \binom{n}{i}t^i(1-t)^{n-i}$ is the Bernstein basis.
The Bézier spline carries an important property: the convex hull property, which guarantees that the trajectory is entirely inside the convex hull of the control points.
Therefore, we can confine the $j$-th piece of trajectory $\boldsymbol{p}_j(t)$ within the spatio-temporal corridors $ \mathcal{P }_j$ by constraining the corresponding control points $\{ \boldsymbol{c}_j^0, \boldsymbol{c}_j^1, \cdots, \boldsymbol{c}_j^n \}$ as follows:
\begin{equation}
	\boldsymbol{A}_{\mathcal{P }_j} \boldsymbol{c}_j^i \leq \boldsymbol{b}_{\mathcal{P }_j},  \quad i \in \{0,1,\cdots,N\},
\end{equation}
By taking the above properties, we can formulate a minimum jerk trajectory optimization problem with spatiotemporal corridor constraints as follows:
\begin{subequations}
	\begin{equation}\label{Eq: cost function}
		\min\limits_{\boldsymbol{c}}  \sum_{j=1}^{T} \int_{t_{j-1}}^{t_j} \left\| \frac{\mathrm{d}^3 \boldsymbol{p}_j(t)}{\mathrm{d} t^3}\right\|^2dt
	\end{equation}
	\begin{equation}\label{Eq: initial condition}
		\text{s.t.} \quad \boldsymbol{p}(t_0)=\boldsymbol{p}_0, \quad \boldsymbol{p}(t_T)=\boldsymbol{p}_T
	\end{equation}
	\begin{equation}\label{Eq: corridor}
		\boldsymbol{p}_j(t)\in \mathcal{P }_j, \forall t \in [t_{j-1}, t_j], ~~ j=1,2,\cdots,T
	\end{equation}
	\begin{equation}\label{Eq: continuity}
		\boldsymbol{p}_{j}(t_{j})=\boldsymbol{p}_{j+1}(t_{j}), ~~ j = 1, 2, \cdots, T
	\end{equation}
	\begin{equation}\label{Eq: vel and scc}
		\boldsymbol{p}^{(1)}_j(t) \leq \boldsymbol{p}^{(1)}_{max}, ~
		\boldsymbol{p}^{(2)}_j(t) \leq \boldsymbol{p}^{(2)}_{max}, ~
		\forall t\in[t_0, t_T], 
	\end{equation}
\end{subequations}
in which $\boldsymbol{p}_0$ and $\boldsymbol{p}_T$ denote the initial and terminal position of the trajectory, and $\boldsymbol{p}^{(1)}_{max}$ and $\boldsymbol{p}^{(2)}_{max}$ give the upper bound of the velocity and acceleration.
Note this is a quadratic programming (QP) problem with linear constraints, which can be solved efficiently by modern QP solvers, e.g. OSQP.

In our method, the time allocation for trajectories is established during the path search phase, as opposed to emerging from the backend optimization.
This strategy is crucial to ensure the safety of the generated trajectories.
As a result, our backend optimization prioritizes spatial optimization with the temporal allocation fixed.
We intentionally avoid a comprehensive spatio-temporal optimization to avoid the computational burden of solving a non-convex optimization problem.
By employing this approach, we not only address the collision avoidance requirements but also achieve a more lightweight backend optimization.

\subsection{Multi-Agent Deconfliction} \label{sec:meth:deconfliction}
The deconfliction is considered to trigger the replanning of each agent.
Due to the Bézier spline's convex hull property, we check the linear separability between the control points of the optimized trajectory and those of the broadcasted trajectory.
Specifically, if the check step identifies a potential collision, the agent will update the SOGM and re-optimize a new trajectory;
Conversely, if no collision is detected, the agent will broadcast and execute the current trajectory.

\section{Results} \label{results}
We compared the performance of the proposed method with MADER \cite{tordesillas2022MADER} and EGO-Swarm \cite{zhou2022Swarm} in simulations of 4 MAVs varying in dynamic obstacles' density and shape.
We then conducted real-world experiments with 2 MAVs in a complex environment with static obstacles and moving pedestrians to demonstrate the effectiveness of our approach.

\subsection{Simulation Experiment}

The 3D simulation experiment was conducted on a laptop equipped with an AMD R7-5800H CPU.
The simulation environment is a 16m$\times$16m$\times$4m 3D space with randomly generated moving columns and circles, implemented based on a quadrotor simulator in \cite{zhou2019Robust}.
Dynamic obstacles are modeled as columns and circles of random sizes, initial positions, and velocities.
The widths of the columns and the radius of the circles are uniformly sampled from 0.5 to 1.0 m and 0.7 to 2.5 m, respectively.
The height of all the columns is 4.0 m, and the width of the circles is fixed at 0.1 m.
The velocities of the obstacles are randomly sampled from 0 to 1.0 m/s, and the directions are uniformly sampled from 0 to 2$\pi$.
Obstacles move with constant velocities and directions, and they rebound when they reach the boundary of the map space.
All MAVs are initialized in the hovering state at a height of 1.0 m, outside the obstacle space.
Their objectives are reaching their individual goals without any collision.

\begin{figure*}[th]
	\centering
	\begin{subfigure}[b]{0.3\textwidth}
		\centering
		\includegraphics[width=0.8\textwidth]{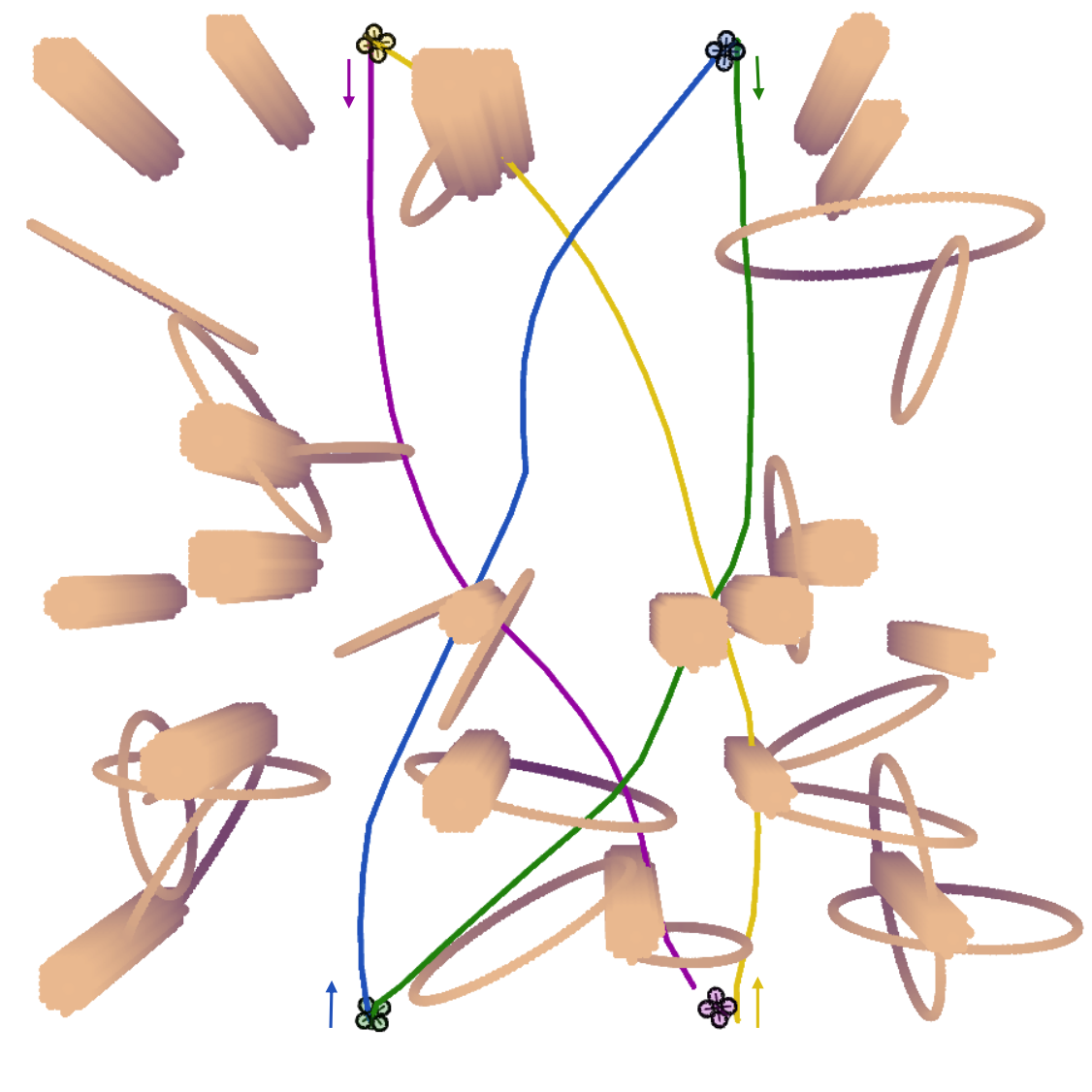}
		\caption{Task 1: Bilateral swap}
		\label{fig:case1}
	\end{subfigure}
	\begin{subfigure}[b]{0.3\textwidth}
            \centering
		\includegraphics[width=0.8\textwidth]{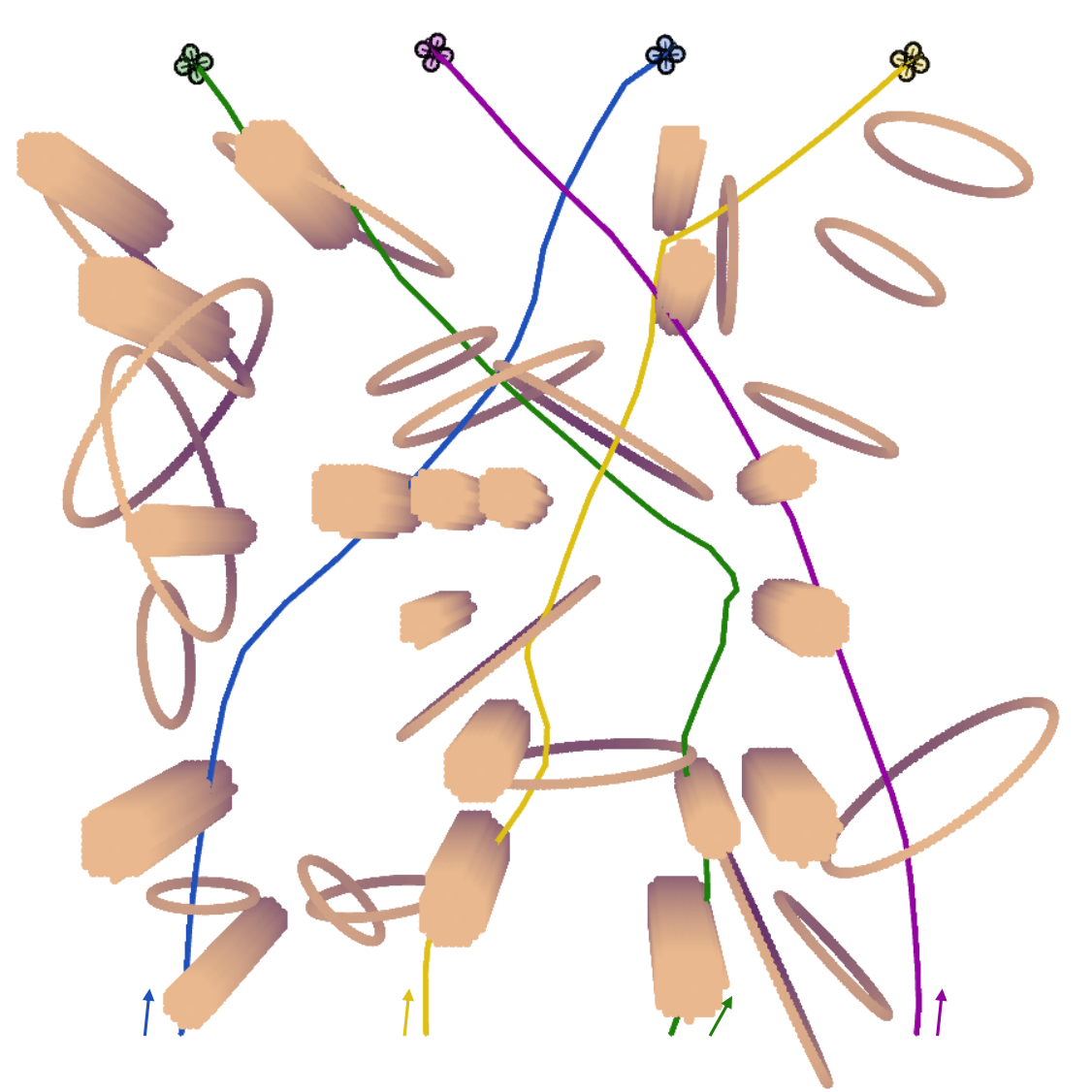}
		\caption{Task 2: Unilateral swap}
		\label{fig:case2}
	\end{subfigure}
	\begin{subfigure}[b]{0.3\textwidth}
            \centering
		\includegraphics[width=0.8\textwidth]{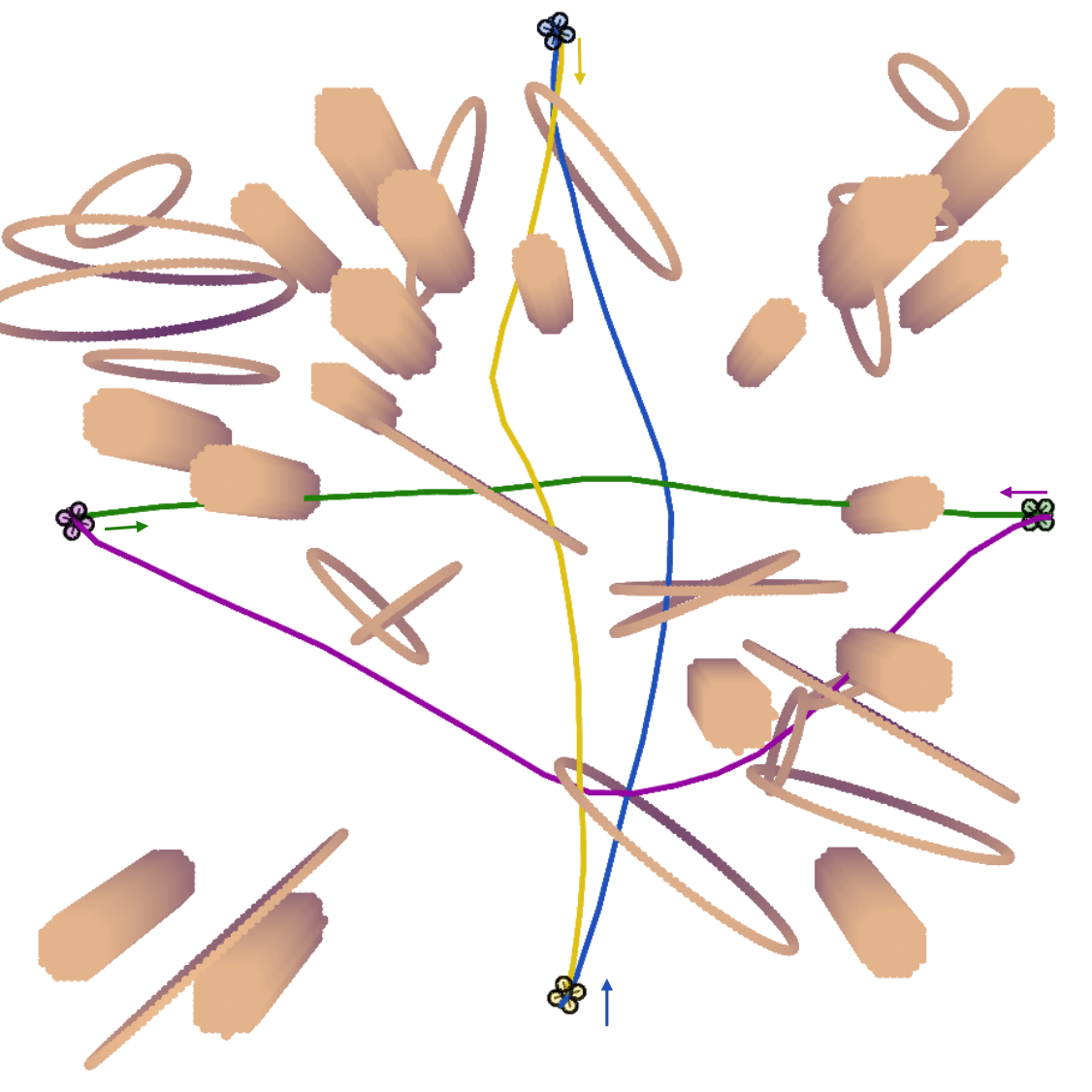}
		\caption{Task 3: Cross swap}
		\label{fig:case3}
	\end{subfigure}
	\vspace*{0.3cm}

	\begin{subfigure}[b]{0.32\textwidth}
		\centering
		\includegraphics[width=\textwidth]{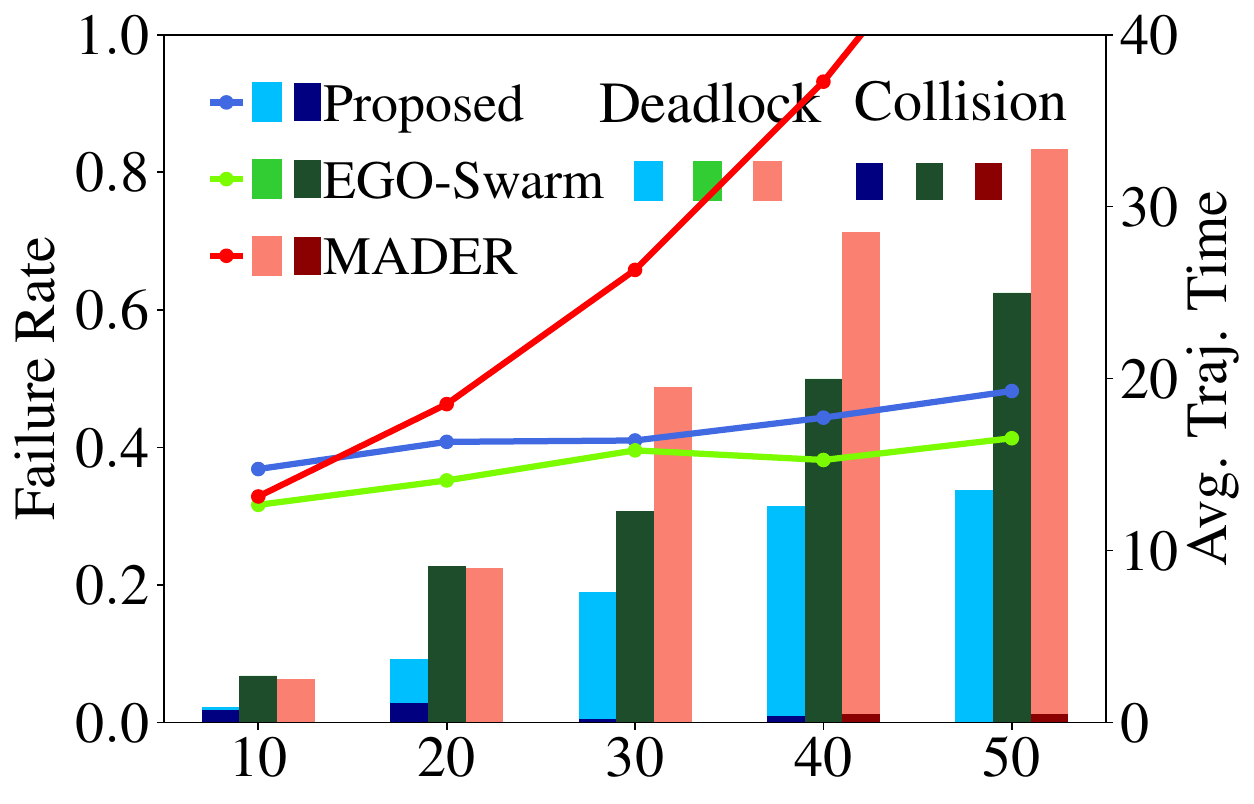}
		\caption{Bilateral swap in mixed environments}
		\label{fig:env1:case1:rst}
	\end{subfigure}
	\begin{subfigure}[b]{0.32\textwidth}
		\includegraphics[width=\textwidth]{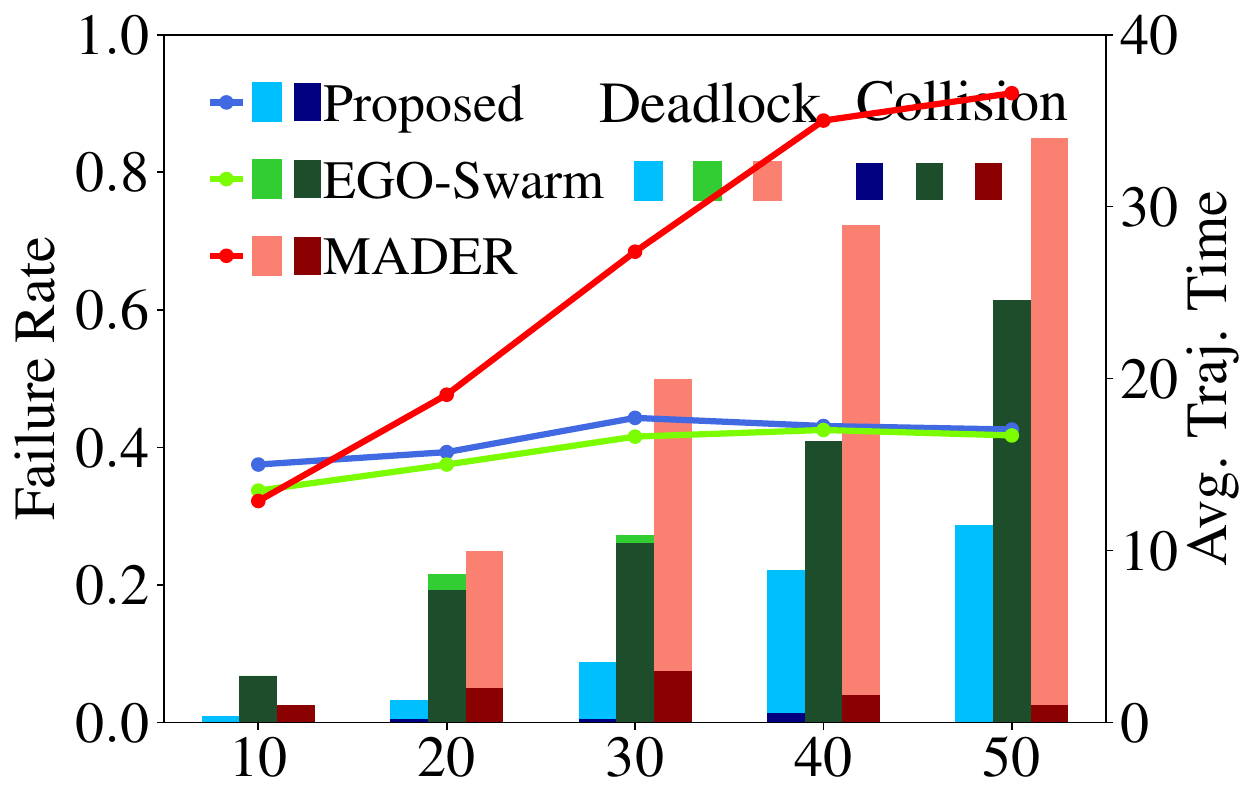}
		\caption{Unilateral swap in mixed environments}
		\label{fig:env1:case2:rst}
	\end{subfigure}
	\begin{subfigure}[b]{0.32\textwidth}
		\includegraphics[width=\textwidth]{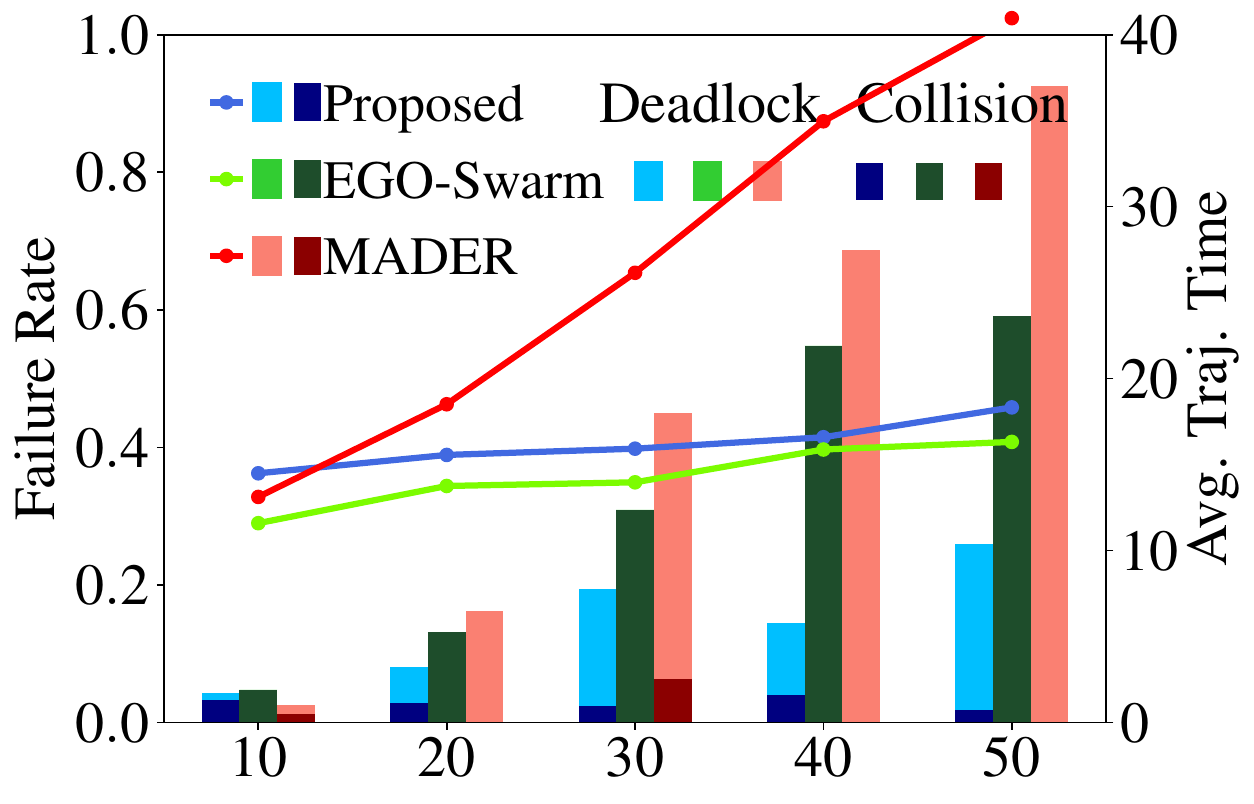}
		\caption{Cross swap in mixed environments}
		\label{fig:env1:case3:rst}
\end{subfigure}
	\vspace*{0.3cm}

	\begin{subfigure}[b]{0.32\textwidth}
		\centering
		\includegraphics[width=\textwidth]{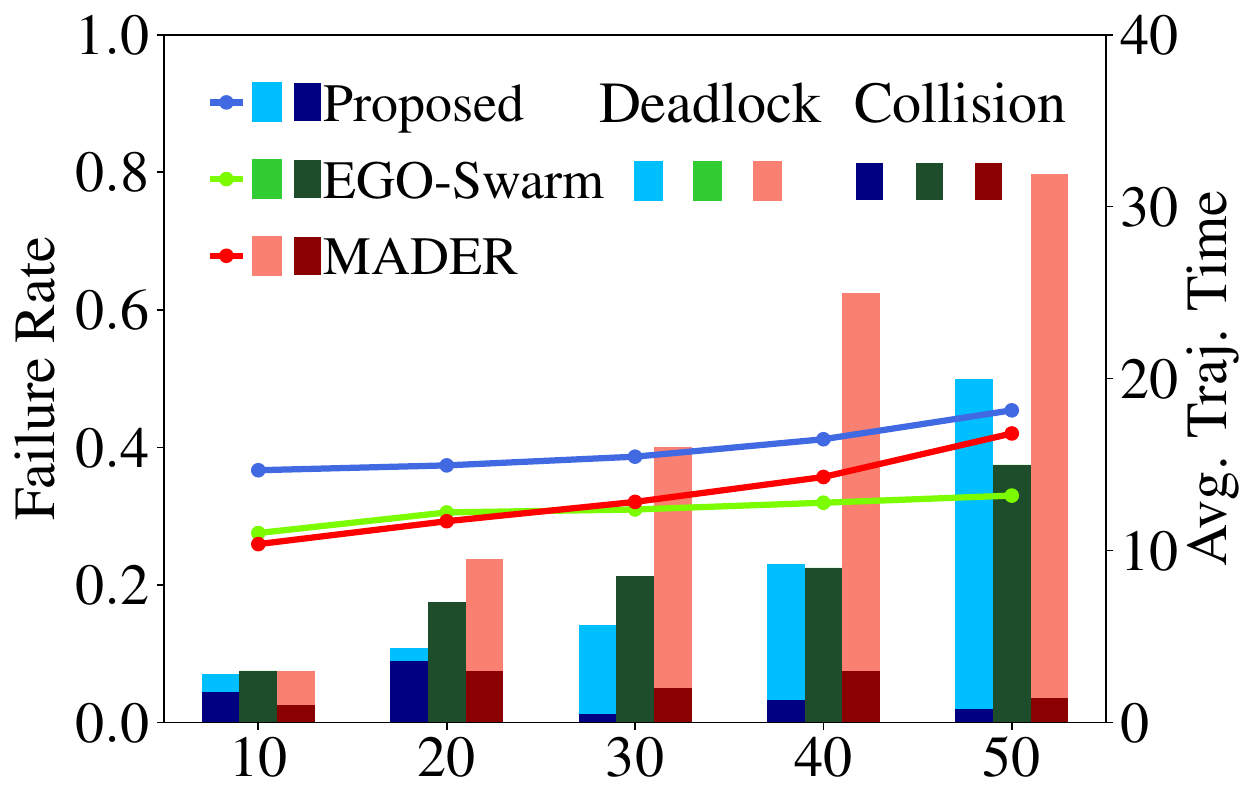}
		\caption{Bilateral swap in pure environments}
		\label{fig:env2:case1:rst}
	\end{subfigure}
	\begin{subfigure}[b]{0.32\textwidth}
		\includegraphics[width=\textwidth]{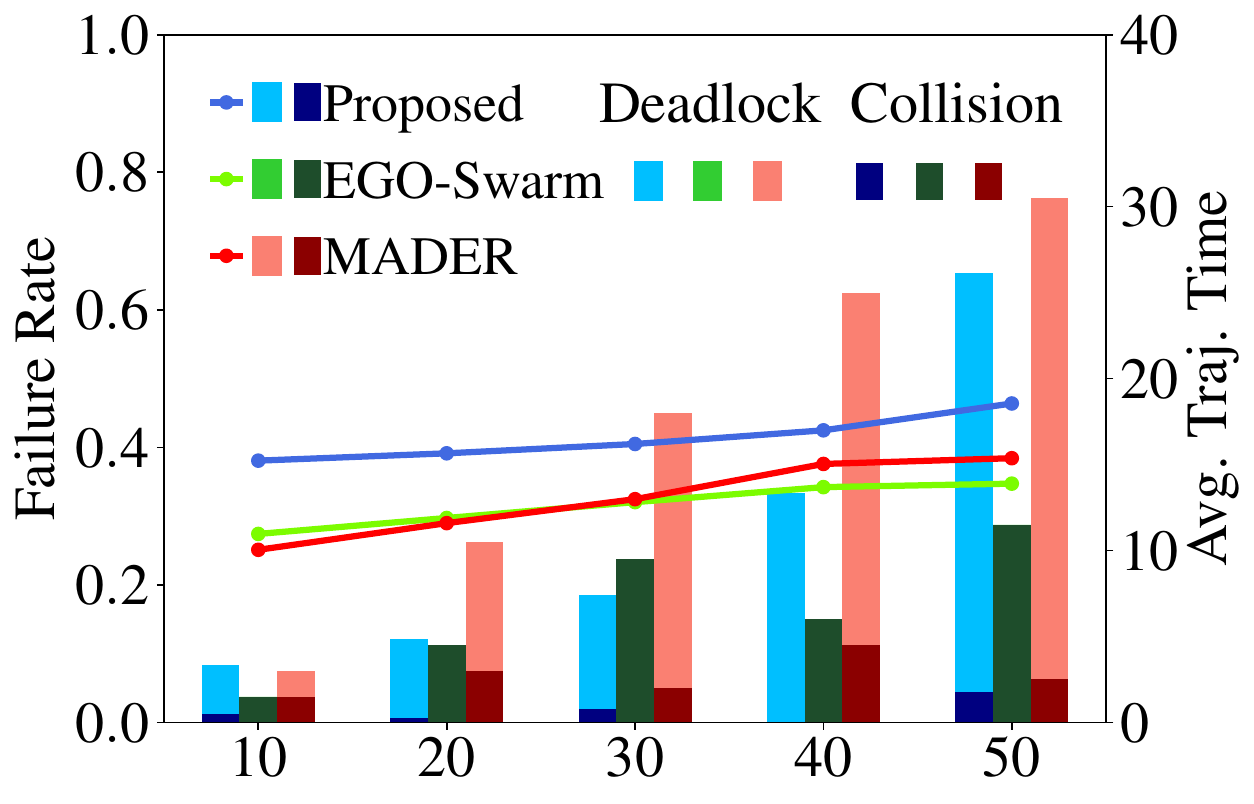}
		\caption{Unilateral swap in pure environments}
		\label{fig:env2:case2:rst}
	\end{subfigure}
	\begin{subfigure}[b]{0.32\textwidth}
		\includegraphics[width=\textwidth]{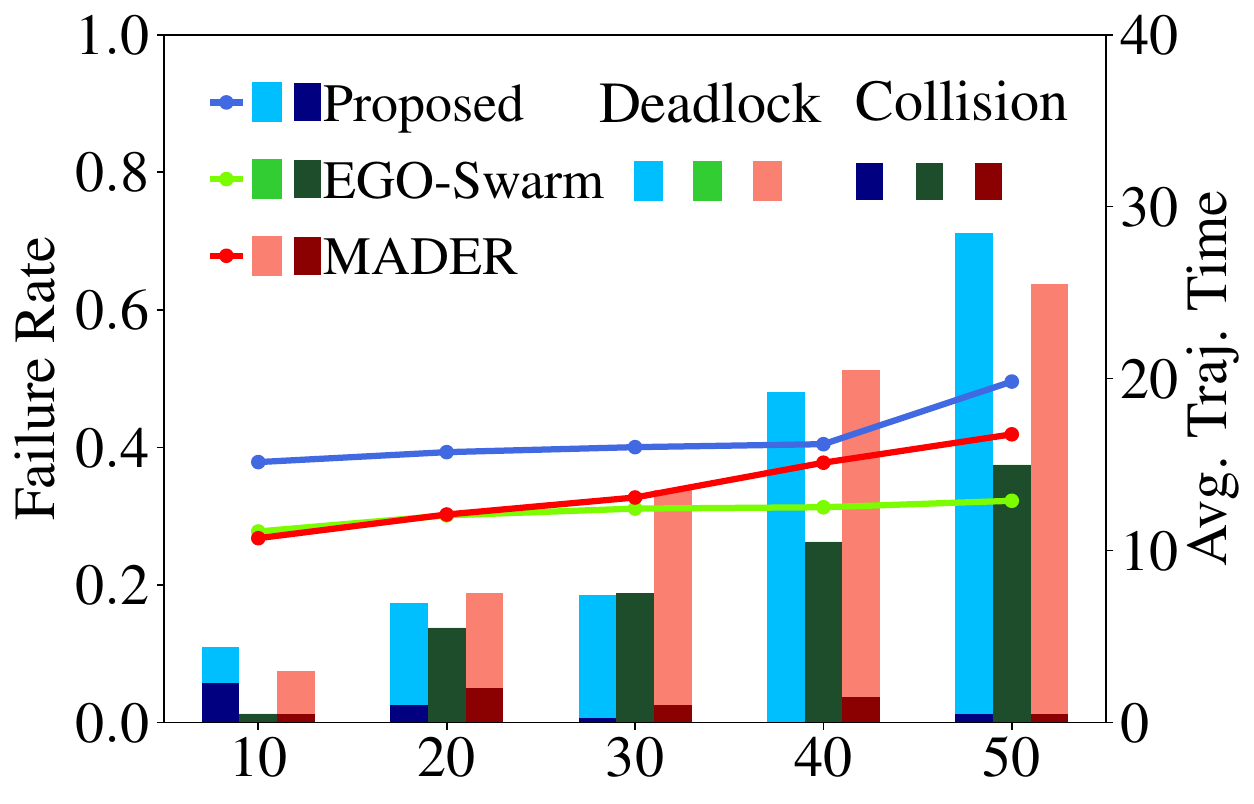}
		\caption{Cross swap in pure environments}
		\label{fig:env2:case3:rst}
	\end{subfigure}

\caption{Performance comparison between our proposed method ({\color{RoyalBlue} Ours}), and two baselines ({\color{BrickRed} MADER} and {\color{ForestGreen} EGO-Swarm}) in 3 different tasks
(Fig.~\ref{fig:case1}--\ref{fig:case3}) and 2 different obstacle settings: mixed environments with both columns and circles (Fig.~\ref{fig:env1:case1:rst}--\ref{fig:env1:case3:rst}),
and pure column environments (Fig.~\ref{fig:env2:case1:rst}--\ref{fig:env2:case3:rst}).
	In Fig.~\ref{fig:case1}--\ref{fig:case3}, trajectories for each MAV are displayed with distinct colors.
Fig.~\ref{fig:env1:case1:rst}--\ref{fig:env2:case3:rst} evaluate the average flight time to the goal and failure rate with different numbers of obstacles,
distinguishing failure types between deadlock and collisions.
	}
	\label{fig:result}
\end{figure*}

The experiments are divided into two types of environments with varying difficulty.
The first type is a \textbf{mixed environment} with columns and circles (Fig.~\ref{fig:env1:case1:rst}--\ref{fig:env1:case3:rst}),
while the second type is a \textbf{pure column environment} (Fig.~\ref{fig:env2:case1:rst}--\ref{fig:env2:case3:rst}) similar to that used in MADER \cite{tordesillas2022MADER}.
Fig.~\ref{fig:case1}--\ref{fig:case3} presents a bird-eye view of these settings in a mixed environment with 20 columns and 20 circles.
This setup aims to create a more cluttered environment by breaking the convexity and isotropy of the obstacles.
Robots can choose either side to avoid columns; however, for circles, they must plan a 3D trajectory to avoid the edges.
Note that robots are allowed to pass through circles from center to center.
To quantify the difficulty, Table \ref{tab:obstacle_density} compares the average obstacle density in both environments that varies in the number of obstacles.
The obstacle density is defined as the ratio of the total volume occupied by obstacles to the volume of the space.

We compare our proposed method against two state-of-the-art baselines, MADER \cite{tordesillas2022MADER} and EGO-Swarm \cite{zhou2022Swarm}, across three distinct position swap tasks: bilateral (Fig.~\ref{fig:case1}), unilateral (Fig.~\ref{fig:case2}), and cross (Fig. \ref{fig:case3} and Fig. \ref{fig:header}). 
Unlike MADER and EGO-Swarm, which necessitate precise shapes and exact trajectories of obstacles, our method employs SOGMs.
To ensure a fair comparison, all methods are confined to using local obstacle information, namely the ground truth positions and velocities within a 5.0 m radius over a time horizon of 2.0 s. 
SOGMs are specifically generated for our approach.
The maximum velocity and acceleration are set to 2.0 m/s and 6.0 m/s$^2$, respectively.
We assume the perfect tracking control in simulation.
More than 50 trials are conducted in each setting. Average results are reported in Fig.~\ref{fig:result}.



\begin{table}[b]
	\centering
	\begin{tabular}{rcccccc}
	\hline
	Num. of Obstacles          & 10  & 20  & 30  & 40  & 50   \\ \hline
	Pure column environment (\%)& 1.73 & 3.45 & 5.18 & 7.00 &  8.62 \\
	Mixed environment  (\%)    &  0.86 & 1.73 & 3.71  & 4.53 & 5.76  \\ \hline
	\end{tabular}
\caption{Average obstacle density in different environments}
	\label{tab:obstacle_density}
\end{table}

The performance is compared in terms of the failure rate and average time to complete the task.
As shown in Fig. \ref{fig:env1:case1:rst}-\ref{fig:env2:case3:rst}, the proposed planner presents 95.3\% and 88.5\% average success rates in the mixed and pure environments when the number of obstacles is less than 30.
This result is comparable to the 87.3\% and 90.8\% success rates of EGO-Swarm, and better than the 87.5\% and 84.8\% success rates of MADER.
In mixed environments with more than 30 obstacles, our method achieves a success rate of 73.0\% between all tasks compared to 45.2\% and 21.0\% for EGO-Swarm and MADER, respectively.
However, in pure environments with more obstacles, our method performs worse than both baselines in success rate and flight time.



To better understand the reason for the failure, we distinguish failures into two different types: collision and deadlock.
Collision only occurs when the agent is following a planned trajectory.
Deadlock is defined as a situation where a dynamic obstacle hits an agent after the agent has stopped due to the inability to find a feasible trajectory.
It reflects the inability to find a feasible trajectory when the search space is exhausted.
In many real-world scenarios, dynamic obstacles such as pedestrians may avoid the robots in a "deadlock" situation.
As shown in Fig. \ref{fig:env1:case1:rst} to Fig. \ref{fig:env1:case3:rst}, deadlock accounts for almost all failures in our method.
In contrast, EGO-Swarm hardly encounters deadlock, but it has a higher collision rate than other methods.
Safety arises because our method confines the planned trajectory within the corridors.
Furthermore, our method performs better in mixed environments than in pure environments.
As depicted in Fig. \ref{fig:env1:case1:rst}-\ref{fig:env2:case1:rst}, a significant decrease in the failure rate can be seen in environments with more obstacles.
These results reveal that the ability of the SOGM to represent obstacles with arbitrary shapes allows for greater flexibility in finding feasible paths in such environments.
Although it is feasible to dissect these obstacles into smaller elements and represent them using AABB or ellipsoidal costs, doing so increases the computational burden, making it challenging to achieve real-time performance.
Therefore, our method is more suitable for such environments. 

\begin{figure}[th]
	\centering
	\includegraphics[width=0.95\linewidth]{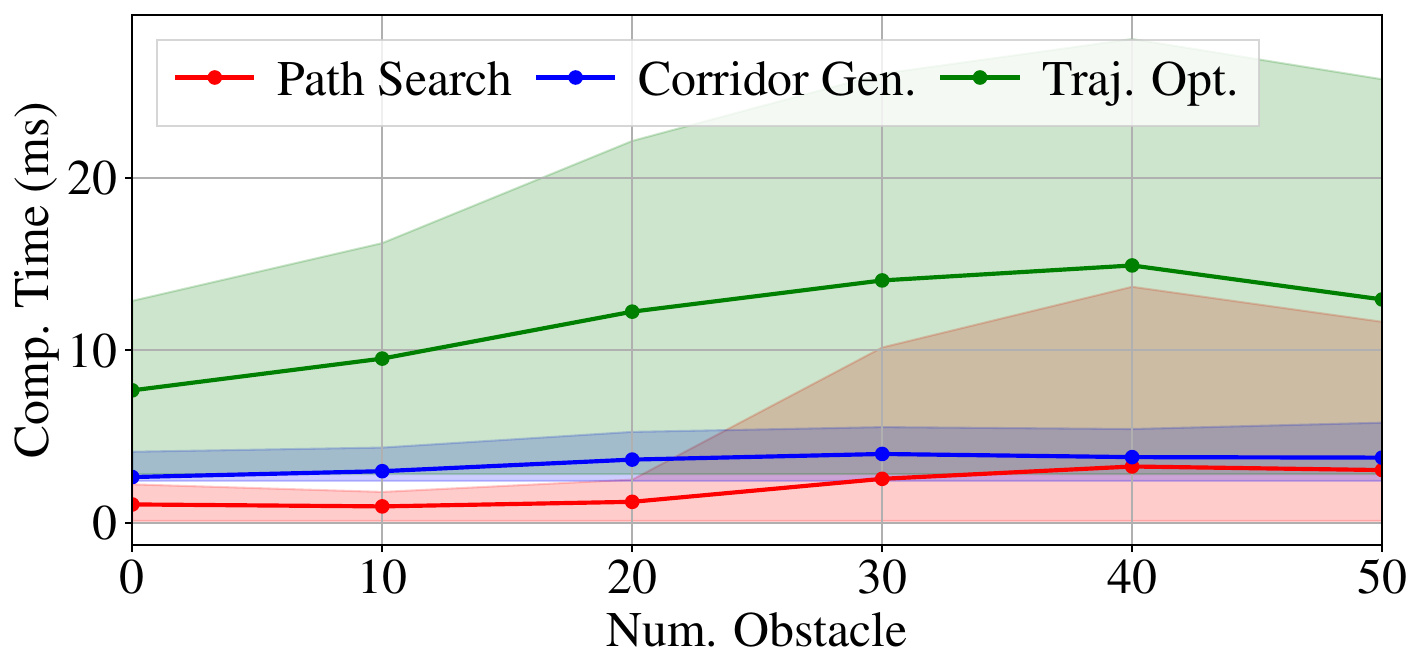}
	\caption{The computation time of each step in our method under different obstacle settings.}
	\label{fig:runtime}
\end{figure}

The run time of each step is evaluated on a laptop with an AMD R7-5800H CPU in mixed environments, shown in Fig. \ref{fig:runtime}.
The overall execution time of our method is 17.19 ms on average, which is significantly faster than MADER \cite{tordesillas2022MADER} (31.04 ms).
The average run time of the kinodynamic path searching step is 1.93 ms, increasing with the number of obstacles.
This is because kinodynamic A* algorithm will traverse more nodes to find a feasible path as the obstacle density increases as in Table \ref{tab:obstacle_density}.
The corridor generation step takes 3.47 ms on average.
The trajectory optimization step takes an average of 11.79 ms, which is time-consuming due to the large number of polygon planes as corridor constraints.
\begin{figure}[th]
	\centering
	\includegraphics[width=1.0\linewidth]{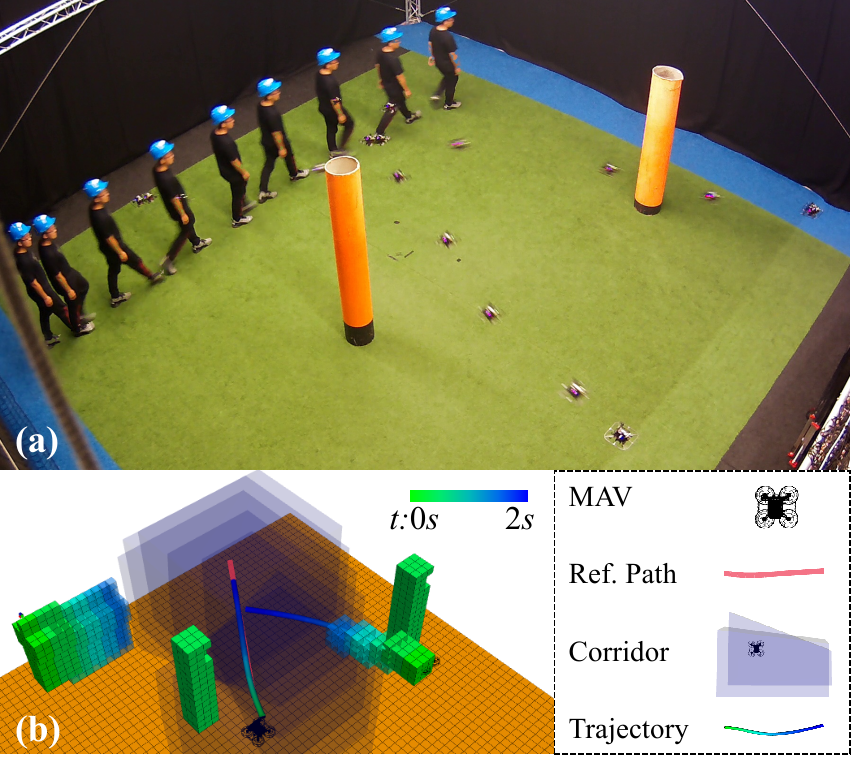}
	\caption{(a) Composite image of two MAVs flying in an indoor dynamic environment.
	(b) The SOGM at $t=2.5s$ is displayed, showing the predicted future occupancies with a gradient transition from green to blue.}
	\label{fig:header}
\end{figure}
\begin{figure}[h]
	\centering
	\includegraphics[width=\linewidth]{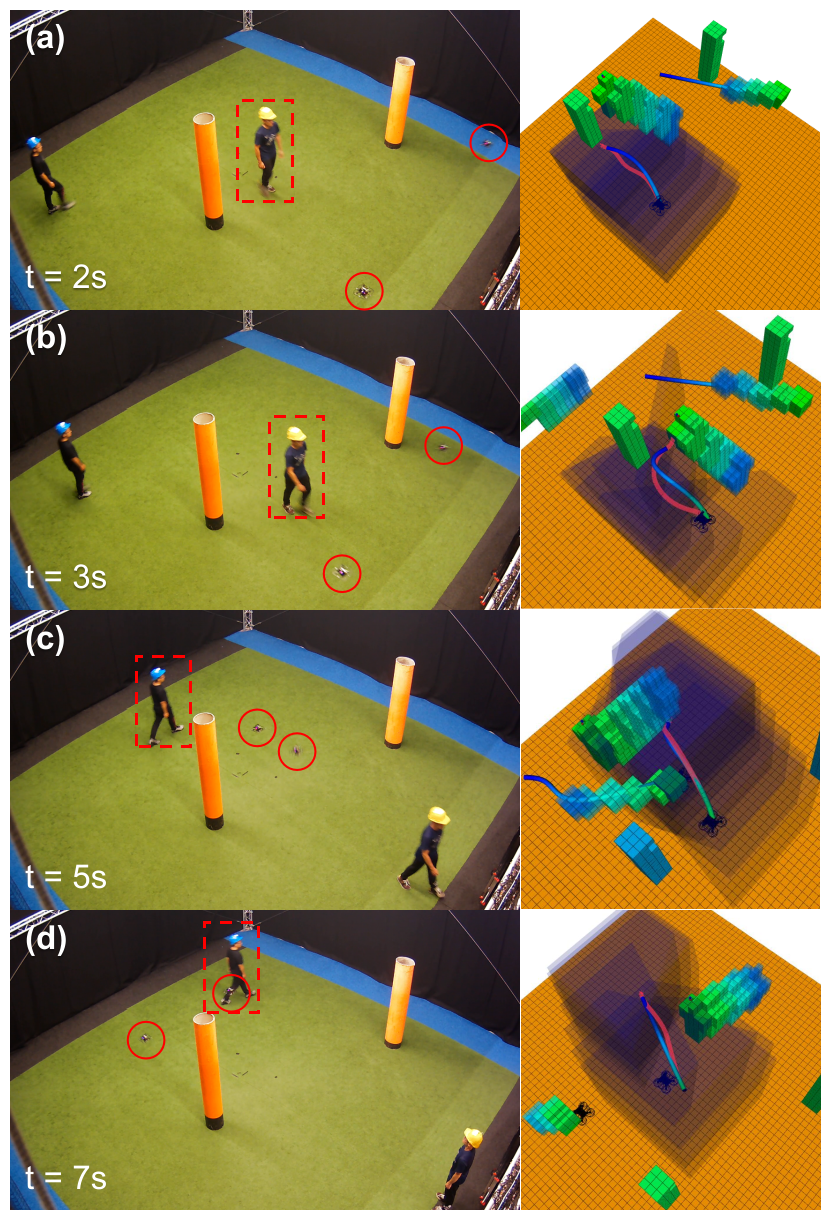}
	\caption{Snapshot of two MAVs flying in a dynamic environment with two pedestrians. Images on the right visualize the SOGM, safety corridors, and planned trajectories.}
	\label{fig:realworld}
\end{figure}
\subsection{Real-World Experiment}
We conducted real-world experiments in a 10m$\times$10m$\times$4m indoor environment to validate the efficiency of the proposed method, shown in Fig. \ref{fig:realworld}.
Each robot is equipped with an NVIDIA Jetson Xavier NX for onboard computation.
Communication between robots is achieved via a WiFi network. 
Considering the limited onboard computational resources, we rely on the OptiTrack system to provide ground-truth information on dynamic obstacles to build the local SOGM.
The robot localization is provided by the OptiTrack as well, due to the limited accuracy and robustness of the onboard localization algorithm in dynamic scenes.
The maximum velocity and acceleration of the robot are set to 1.0 m/s and 4.0 m/s$^2$.
The planner runs at 10 Hz.
Experiment settings and the planning results are visualized in Fig.~\ref{fig:header} and Fig.~\ref{fig:realworld}.
For more details, please refer to the supplementary video.

\section{Conclusion} \label{conclusion}
This paper presents a decentralized multi-agent trajectory planning framework that is able to handle environments with arbitrarily shaped static and dynamic obstacles.
Our planner adopts the kinodynamic A* path-searching algorithm on the SOGM, allowing the robots to find a feasible path in such environments.
The corridor generation and trajectory optimization methods efficiently help generate safe and dynamically feasible trajectories.
Simulation results show that our method achieves more than 88.5\% success rate in dynamic environments with less than 20 obstacles, which is comparable to both baselines.
In dynamic environments where obstacles cannot be represented by AABB or ellipsoids, our method outperforms the baselines by achieving a 51.9\% and 27.7\% lower failure rate than MADER and EGO-Swarm, respectively.
By classifying the failure cases into collision and deadlock, we find that the majority of our failures are due to deadlock cases.
Real-world experiments further validate the effectiveness of the proposed method.
However, in environments dense with obstacles or when there are numerous robots, this method is susceptible to significant deadlock issues.
This can be attributed to the absence of global guidance and the loss of topological homotopy under frequent replanning.
In future research, we will focus on multi-agent path finding with topological homotopy to mitigate these deadlock issues.

\bibliographystyle{IEEEtran}
\bibliography{references}

\begin{thebibliography}{10}
\providecommand{\url}[1]{#1}
\csname url@rmstyle\endcsname
\providecommand{\newblock}{\relax}
\providecommand{\bibinfo}[2]{#2}
\providecommand\BIBentrySTDinterwordspacing{\spaceskip=0pt\relax}
\providecommand\BIBentryALTinterwordstretchfactor{4}
\providecommand\BIBentryALTinterwordspacing{\spaceskip=\fontdimen2\font plus
\BIBentryALTinterwordstretchfactor\fontdimen3\font minus
  \fontdimen4\font\relax}
\providecommand\BIBforeignlanguage[2]{{%
\expandafter\ifx\csname l@#1\endcsname\relax
\typeout{** WARNING: IEEEtran.bst: No hyphenation pattern has been}%
\typeout{** loaded for the language `#1'. Using the pattern for}%
\typeout{** the default language instead.}%
\else
\language=\csname l@#1\endcsname
\fi
#2}}

\bibitem{wang2021Autonomous}
Y.~Wang, J.~Ji, Q.~Wang, C.~Xu, and F.~Gao, ``Autonomous {{Flights}} in
  {{Dynamic Environments}} with {{Onboard Vision}},'' in \emph{2021
  {{IEEE}}/{{RSJ International Conference}} on {{Intelligent Robots}} and
  {{Systems}} ({{IROS}})}, 2021, pp. 1966--1973.

\bibitem{9359513}
G.~Chen, W.~Dong, X.~Sheng, X.~Zhu, and H.~Ding, ``An active sense and avoid
  system for flying robots in dynamic environments,'' \emph{IEEE/ASME
  Transactions on Mechatronics}, vol.~26, no.~2, pp. 668--678, 2021.

\bibitem{zhou2022Swarm}
X.~Zhou, X.~Wen, Z.~Wang, Y.~Gao, H.~Li, Q.~Wang, T.~Yang, H.~Lu, Y.~Cao,
  C.~Xu, and F.~Gao, ``Swarm of micro flying robots in the wild,''
  \emph{Science Robotics}, vol.~7, no.~66, p. eabm5954, 2022/05/11.

\bibitem{zhu2019ChanceConstrained}
H.~Zhu and J.~Alonso-Mora, ``Chance-{{Constrained Collision Avoidance}} for
  {{MAVs}} in {{Dynamic Environments}},'' \emph{{IEEE} Robot. Autom. Lett.
  (RA-L)}, vol.~4, no.~2, pp. 776--783, 2019-04.

\bibitem{xu2022DPMPCPlanner}
Z.~Xu, D.~Deng, Y.~Dong, and K.~Shimada, ``{{DPMPC-Planner}}: {{A}} real-time
  {{UAV}} trajectory planning framework for complex static environments with
  dynamic obstacles,'' in \emph{2022 {{Int}}. {{Conf}}. {{Robot}}. {{Autom}}.
  {{ICRA}}}, 2022, pp. 250--256.

\bibitem{tordesillas2022MADER}
J.~Tordesillas and J.~P.~How, ``Mader: Trajectory planner in multiagent and
  dynamic environments,'' \emph{{IEEE} Trans. Robot. (T-RO)}, vol.~38, no.~1,
  pp. 463--476, 2022.

\bibitem{Learning2022Tomas}
H.~Thomas, M.~G. de~Saint~Aurin, J.~Zhang, and T.~D. Barfoot, ``Learning
  spatiotemporal occupancy grid maps for lifelong navigation in dynamic
  scenes,'' in \emph{2022 Intl. Conf. on Robot. and Autom. (ICRA)}, 2022, pp.
  484--490.

\bibitem{OursMap}
G.~Chen, W.~Dong, P.~Peng, J.~Alonso-Mora, and X.~Zhu, ``Continuous occupancy
  mapping in dynamic environments using particles,'' \emph{arXiv preprint
  arXiv:2202.06273}, 2022.

\bibitem{hou2022Enhanced}
J.~Hou, X.~Zhou, Z.~Gan, and F.~Gao, ``Enhanced decentralized autonomous aerial
  robot teams with group planning,'' \emph{IEEE Robot. Autom. Lett. (RA-L)},
  vol.~7, no.~4, pp. 9240--9247, 2022.

\bibitem{hornung2013octomap}
A.~Hornung, K.~M. Wurm, M.~Bennewitz, C.~Stachniss, and W.~Burgard, ``Octomap:
  An efficient probabilistic 3d mapping framework based on octrees,''
  \emph{Auton. Robots}, vol.~34, no.~3, pp. 189--206, 2013.

\bibitem{Esperance2014Safety}
B.~L’Espérance and K.~Gupta, ``Safety hierarchy for planning with time
  constraints in unknown dynamic environments,'' \emph{IEEE Transactions on
  Robotics}, vol.~30, no.~6, pp. 1398--1411, 2014.

\bibitem{lavalle2006Planning}
S.~M. LaValle, \emph{Planning {{Algorithms}}}.\hskip 1em plus 0.5em minus
  0.4em\relax {Cambridge University Press}, 2006.

\bibitem{lin2020Robust}
J.~Lin, H.~Zhu, and J.~Alonso-Mora, ``Robust {{Vision-based Obstacle
  Avoidance}} for {{Micro Aerial Vehicles}} in {{Dynamic Environments}},'' in
  \emph{2020 Intl. Conf. on Robot. and Autom. ({{ICRA}})}, 2020, pp.
  2682--2688.

\bibitem{Mina2017Robust}
M.~Kamel, J.~Alonso-Mora, R.~Siegwart, and J.~Nieto, ``Robust collision
  avoidance for multiple micro aerial vehicles using nonlinear model predictive
  control,'' in \emph{2017 IEEE/RSJ Intl. Conf. on Intell. Robots and Syst.
  (IROS)}, 2017, pp. 236--243.

\bibitem{gao2017Quadrotor}
F.~Gao and S.~Shen, ``Quadrotor trajectory generation in dynamic environments
  using semi-definite relaxation on nonconvex {{QCQP}},'' in \emph{2017 Intl.
  Conf. on Robot. and Autom. ({{ICRA}})}, 2017-05, pp. 6354--6361.

\bibitem{chen2022Realtime}
H.~Chen and P.~Lu, ``\BIBforeignlanguage{en}{Real-time identification and
  avoidance of simultaneous static and dynamic obstacles on point cloud for
  {{UAVs}} navigation},'' \emph{\BIBforeignlanguage{en}{Robotics and Autonomous
  Systems}}, vol. 154, p. 104124, 2022.

\bibitem{10034468}
G.~Chen, P.~Peng, P.~Zhang, and W.~Dong, ``Risk-aware trajectory sampling for
  quadrotor obstacle avoidance in dynamic environments,'' \emph{IEEE
  Transactions on Industrial Electronics}, pp. 1--10, 2023.

\bibitem{chen2023RAST}
G.~Chen, S.~Wu, M.~Shi, W.~Dong, H.~Zhu, and J.~Alonso-Mora, ``{{RAST}}:
  {{Risk-Aware Spatio-Temporal Safety Corridors}} for {{MAV Navigation}} in
  {{Dynamic Uncertain Environments}},'' \emph{IEEE Robot. Autom. Lett. (RA-L)},
  vol.~8, no.~2, pp. 808--815, 2023.

\bibitem{honig2018Trajectory}
W.~H\"onig, J.~A. Preiss, T.~K.~S. Kumar, G.~S. Sukhatme, and N.~Ayanian,
  ``Trajectory {{Planning}} for {{Quadrotor Swarms}},'' \emph{{IEEE} Trans.
  Robot. (T-RO)}, vol.~34, no.~4, pp. 856--869, 2018-08.

\bibitem{zhu2019BUAVC}
H.~Zhu and J.~Alonso-Mora, ``B-{{UAVC}}: {{Buffered Uncertainty-Aware Voronoi
  Cells}} for {{Probabilistic Multi-Robot Collision Avoidance}},'' in
  \emph{2019 {{International Symposium}} on {{Multi-Robot}} and {{Multi-Agent
  Systems}} ({{MRS}})}, 2019, pp. 162--168.

\bibitem{luis2020Online}
C.~E. Luis, M.~Vukosavljev, and A.~P. Schoellig, ``Online {{Trajectory
  Generation With Distributed Model Predictive Control}} for {{Multi-Robot
  Motion Planning}},'' \emph{{IEEE} Robot. Autom. Lett. (RA-L)}, vol.~5, no.~2,
  pp. 604--611, 2020.

\bibitem{kondo2023Robust}
K.~Kondo, J.~Tordesillas, R.~Figueroa, J.~Rached, J.~Merkel, P.~C. Lusk, and
  J.~P. How, ``Robust mader: Decentralized and asynchronous multiagent
  trajectory planner robust to communication delay,'' in \emph{2023 IEEE Intl.
  Conf. on Robot. and Autom. (ICRA)}, 2023, pp. 1687--1693.

\bibitem{park2021Online}
J.~Park and H.~J. Kim, ``Online {{Trajectory Planning}} for {{Multiple
  Quadrotors}} in {{Dynamic Environments Using Relative Safe Flight
  Corridor}},'' \emph{{IEEE} Robot. Autom. Lett. (RA-L)}, vol.~6, no.~2, pp.
  659--666, 2021.

\bibitem{mann2022Predicting}
K.~S. Mann, A.~Tomy, A.~Paigwar, A.~Renzaglia, and C.~Laugier, ``Predicting
  {{Future Occupancy Grids}} in {{Dynamic Environment}} with {{Spatio-Temporal
  Learning}},'' 2022.

\bibitem{zhou2019Robust}
B.~Zhou, F.~Gao, L.~Wang, C.~Liu, and S.~Shen, ``Robust and {{Efficient
  Quadrotor Trajectory Generation}} for {{Fast Autonomous Flight}},''
  \emph{{IEEE} Robot. Autom. Lett. (RA-L)}, vol.~4, no.~4, pp. 3529--3536,
  2019-10.

\bibitem{liu2017Planning}
S.~Liu, M.~Watterson, K.~Mohta, K.~Sun, S.~Bhattacharya, C.~J. Taylor, and
  V.~Kumar, ``Planning {{Dynamically Feasible Trajectories}} for {{Quadrotors
  Using Safe Flight Corridors}} in 3-{{D Complex Environments}},'' \emph{{IEEE}
  Robot. Autom. Lett. (RA-L)}, vol.~2, no.~3, pp. 1688--1695, 2017.

\bibitem{wang2022Geometrically}
Z.~Wang, X.~Zhou, C.~Xu, and F.~Gao, ``Geometrically {{Constrained Trajectory
  Optimization}} for {{Multicopters}},'' \emph{{IEEE} Trans. Robot. (T-RO)},
  pp. 1--10, 2022.

\end{thebibliography}

\end{document}